# Arabic Fake News Detection Based on Deep Contextualized Embedding Models

[1,2]Ali Bou Nassif*, [3]Ashraf Elnagar, [1]Omar Elgendy, [1]Yaman Afadar
[1]Department of Computer Engineering, University of Sharjah, Sharjah, UAE, P.O. Box: 27272
[2]Western University, London, ON, Canada, N6A 3K7
[3]Department of Computer Science, University of Sharjah, Sharjah, UAE, P.O. Box: 27272
{anassif, ashraf, u20106090, u17104387}@sharjah.ac.ae

*Corresponding Author (Ali Bou Nassif, anassif@sharjah.ac.ae)

*Abstract*—social media is becoming a source of news for many people due to its ease and freedom of use. As a result, fake news has been spreading quickly and easily regardless of its credibility, especially in the last decade. Fake news publishers take advantage of critical situations such as the Covid-19 pandemic and the American presidential elections to affect societies negatively. Fake news can seriously impact society in many fields including politics, finance, sports, etc. Many studies have been conducted to help detect fake news in English, but research conducted on fake news detection in the Arabic language is scarce. Our contribution is twofold: first, we have constructed a large and diverse Arabic fake news dataset. Second, we have developed and evaluated transformer-based classifiers to identify fake news while utilizing eight state-of-the-art Arabic contextualized embedding models. The majority of these models had not been previously used for Arabic fake news detection. We conduct a thorough analysis of the state-of-the-art Arabic contextualized embedding models as well as comparison with similar fake news detection systems. Experimental results confirm that these state-of-the-art models are robust, with accuracy exceeding 98%.

*Keywords— Arabic fake news; Natural language processing; Contextualized Models; Deep learning*

## I. Introduction

According to a survey conducted in December of 2016, 64% of Americans are confused because of the fake news they see every day; 24% think that fake news makes them a little confused, and the remaining 11% do not care about the spreading of fake news and think it is not confusing to them.[1] This shows us the importance of coming up with an accurate and efficient solution to eliminate the spread of fake news.

---

[1] https://www.journalism.org/2016/12/15/many-americans-believe-fake-news-is-sowing-confusion/pj_2016-12-15_fake-news_0-01/.

Nowadays, accessing social media has become very simple. It is a major source of the news consumed each day by many people [1]. For instance, on Instagram, more than 995 photos are posted per second. On Facebook, more than 4000 new posts appear each second globally.[2] This makes it so easy for anyone to write, share and publish news about any field. Unfortunately, accessing social media easily and freely has disadvantages [2]. Many users share fake and misleading news items, which has a negative impact on society [3, 4].

Machine learning algorithms including classical and deep models have been used in several disciplines such as Speech recognition, Natural Language Processing, Security, etc. [5][6]. Sharing fake news is becoming increasingly common. Fake news is defined as false news or news containing misleading information. Fake news has become a prevalent issue in many fields, especially about financial and political issues, and detecting untrustworthy news sources has become very difficult with the high amount of traffic.

As UCF NEWS recently published, fake news can have a disproportionate negative impact on events such as the US elections. The creators of fake news come up with clear stories in a way that the human brain cannot detect as fake.[3] Falsified news items also play an important role in critical events such as the spread of Covid-19, which has been recognized as a global pandemic [7, 8]. In the Arab world, fake news has affected events taking place throughout Arabic-speaking countries. Uncertain times become even harder when members of the public are not able to determine what events truly took place and what is being reported falsely [9].

A lot of previous work was conducted on detecting fake news using deep learning pre-trained models from transformers. However, most of this work was focused on fake news written in English. As a result, publications addressing fake news detection on Arabic-language social media forums are very rare. The reason is that Arabic natural language processing is one of the most challenging fields in Natural language processing. First, the form and the spelling of the words can differ from one sentence to another, and this difference can completely change the meaning of the word. For example, the word "ذهب" as a verb means "Go" and as a noun means "Gold". It can be identified only through the sentence and the pronunciation. In addition, Arabic language has many dialects, and some dialects are completely different from other dialects[4].

In this work, we make important contributions to the field of knowledge. This includes:

---

[2] https://www.omnicoreagency.com/
[3] https://www.ucf.edu/news/how-fake-news-affects-u-s-elections/
[4] https://towardsdatascience.com/arabic-nlp-unique-challenges-and-their-solutions-d99e8a87893d

- Due to the lack of Arabic language fake news detection datasets, we collected our own data using web scraping and official Arabic news data. Detailed procedure is explained in the Methodology section.

- We translated English fake news detection from Kaggle into Arabic using python code.

- We used the above-mentioned datasets to compare the performance. The originally Arabic dataset that we construct performs better using the BERT model for Arabic text (ex: AraBert, QaribBert, etc.)

- We compared eight deep learning classifiers using state-of-the-art Arabic contextualized embedding models on the collected datasets. We then assessed the performance of all the models, with an in-depth discussion of false negative and false positive results.

- Many of the used models in this paper have not been used before in the Arabic language fake news detection and we obtained a new state-of-art result using AraBert and QaribBert.

This paper is structured as follows: In the next section we discuss related work. In Section III, we set out the methodology we used, including an overview of the datasets, the preprocessing techniques, the models and their configuration. In Section IV, our experimental approach is discussed in detail and each model's performance with the two datasets is evaluated according to accuracy, precision, recall and F-score. Finally, we conclude our work in Section V with a summary and a discussion of limitations and future work.

## II. RELATED WORK

Natural Language Processing (NLP) for Arabic language has become a very interesting and challenging topic for researchers with its various topics and tasks [10]. In addition to the fake news detection and spam detection, there are many important and related tasks to begin with such as Arabic sentiment analysis (ASA), question answering system in Arabic language, etc.

Several literature reviews have been conducted on the ASA to explain the must use methodology for the recent work, the research gaps and challenges. The reviews [11] [12] [13] covered the topic ASA in detail and provided a good reference for the researchers to start from.

Starting with the comprehensive review by [11], the authors stated all the methods, steps and challenges that face researchers in the Arabic sentiment analysis (ASA). They selected more than 100 papers dated from 2006 to 2019.

The authors listed all the Arabic corpora designed for sentiment analysis and compare the recent works performance. According to their results, the most used classifiers were Support Vector Machine (SVM), K-nearest neighbors (KNN) and Naïve Bayes (NB). Lastly, many pre-processing and feature extraction methods have been discussed.

Similar to the above-mentioned review, the systematic review by [12] also discussed ASA task and proposed new research avenues in this area. The review article includes 71 papers published between January 2000 and June 2020. The authors classified the publications based on many categorization criteria. The result showed that 25% of the papers have used Convolution Neural Network (CNN) to do ASA, and 20% have used Long Short-Term Memory (LSTM). The review also covered the most popular data sources for ASA task and elaborate on each.

The review article by [13] summarized a comprehensive review of 60 research conducted on the Arabic dialect sentiment analysis published between 2012 and 2020. As stated on the above-mentioned review, their findings revealed that SVM was the most used machine learning algorithm. The next is NB and from deep learning, the LSTM model adopted around 9% of the overall papers. The result also pointed that Saudi, Egyptian, Jordanian, and Algerian dialects were found to be the most studied.

On the sentiment analysis for multilanguage, The publication by [14] proposed SenticNet versions for 40 languages. The result showed that 30% to 60% of the semantics associated with the idea are correct based on the target language. The authors also stated that their method is low-cost and not time consuming.

The review written by [15] discussed Arabic language processing (ALP) and modeling techniques for question answering system. The authors also discussed all the existing methods, datasets and processing techniques for the task, and elaborated on the challenges and limitations.

Most recent published work about fake news detection has focused on English-language news. Very few studies have been done on Arabic sources and those in other languages. In the following paragraph we grouped the recent work on the fake Arabic news detection.

Authors in [16] mentioned the lack of sufficient data for Arabic fake news detection. They proposed to conduct their research on only the true stories that are abundantly available online, using a part of speech tagger (POS). The authors used different BERT-based models (mBert, XLMR, Arabert) and achieved powerful performance using

Arabert. They evaluated their models according to accuracy and F-score, which were recorded at 89.23% and 89.25% respectively. Also, they mentioned that future work would involve collecting datasets from AraNews using stories generated by AraNews. Similarly, the authors in [17] conducted several exploratory analyses in order to identify the linguistic properties of Arabic fake news but with sarcastic content. Their main contribution was to show that although news items with sarcastic content might be trustworthy and verifiable, the text has distinguishing features on the lexica-grammatical level. They built several machine-learning models to identify sarcastic fake news and achieved accuracy up to 98.6% using two Arabic datasets collected from sarcastic news websites that contain only fake news. Their real news dataset was sourced from BBC-Arabic and CNN-Arabic. The content in their datasets was mostly about political issues in the Middle East.

While the first two articles proposed supervised classifiers, authors in [18] focused on analyzing the credibility of online blogs semi-supervised end-to-end deep learning. They found that the lack of availability of Arabic datasets was a difficulty. To overcome this issue, they proposed a deep co-learning model to assess the credibility of Arabic blogs. In this model, they trained multiple weak deep neural networks on a small, labeled dataset, with each network using a different view of the data. Each one of these classifiers was then used to classify unlabeled data. Its predictions were used to train the other classifiers in a form of semi-supervised learning. Comparing different methods such as SVM and CNN, they reported a promising performance. Their method achieved a top F-score of 63%.

To improve Arabic fake news detection, authors in [1] utilized content and user-related features to apply sentiment analysis to generate new features that would improve the binary classification performance. They tested their features and built four new machine-learning models using Random Forest, Decision Tree, AdaBoost and Logistic Regression. They used a dataset that contained 2708 tweets, which was filtered down to 1862 balanced tweets published on topics covering the Syrian crisis. They achieved a strong performance measured using accuracy, precision, recall and F-score, which were recorded at 74%, 78%, 80% and 79% respectively.

Authors in [19], addressed the detection of untrustworthy sources using low dimensionality statistical embedding. They used this method on several Arabic-language corpora including the Arabic credibility corpus and two other corpora they created from Qatar News. The methods they used produced an F-score of 79.7%. Their results were obtained with two well-known distributed representations, namely Continuous Bag of Words and Skip Grams.

Another work that is related to ours attempted to perform an extensive analysis on the credibility of Arabic content on Twitter [20]. The authors built a classification model that could predict the credibility of a given Arabic tweet. They used a dataset composed of 9000 Arabic tweets independent of the topic, extracting different features about author profile and timeline. They observed a very strong performance with this model, registering a 21% improvement from the standard F-score at that time. In addition, they conducted an experiment to highlight the importance of user-based features and content-based features.

Furthermore, the authors in [21], addressed the issue of finding evidence for Arabic news claims. They conducted four tasks using machine learning, using a different set of features in each task. They used three machine-learning models: Naïve Bayes, support vector machine and random forest. Using different models with various features, a F-score of 83% was conserved.

Different from the above-mentioned publications, the authors in [22], used a hybrid of topic and user features to evaluate news credibility. The authors proposed a machine-learning model to classify fake news on Twitter. The proposed model consists of four main modules: a) content parsing and features extraction, b) content verification, c) user comments polarity evaluation and d) credibility classification. A dataset of 800 manually labeled Arabic news items was collected from Twitter. They used three models: decision tree, support vector machine and Naive Bayes. Results indicate that the decision tree model achieved a true positive rate that was around 2% higher than the support vector machine model and 7% higher than the Naïve Bayes model. Furthermore, the false positive rate produced by the decision tree model was almost 9% lower than that of the support vector machine model and 10% lower than Naive Bayes. For precision, recall, F-score and accuracy, the decision tree model achieved almost 2% higher results than the support vector machine model and 7% higher than the Naïve Bayes model on the tested dataset.

In [23], the authors conducted a benchmark study to assess the performance of different approaches using three datasets. The authors developed the largest and most diversified dataset ever assembled. In addition, they developed some advanced deep learning models that performed very well. The best performance was achieved using Naïve Bayes when the two datasets were combined, with accuracy, precision, recall and F-score recorded at 95%.

Both authors [24] and [25] discussed Arabic fake news detection methods and results. In [24], article have used machine learning based models to develop fake news detection classifier. They used YouTube API to collect their data and retrieve the related comments. Their model obtained an accuracy of 95% using SVM algorithm. Whereas authors in [25] used covid-19 hashtags to collect Covid-19 related tweets. They used six classifiers to do the task

(Naïve Bayes, Logistic Regression, Support Vector Machine, Multilayer Perceptron, Random Forest Bagging, and eXtreme Gradient Boosting) and they achieved the highest F1-score of 93.3% using LR method.

The review article by [26] presented an overall review for Arabic fake news detection task and compare between various linear and deep learning models CNN, Recurrent Neural Network (RNN), Gated recurrent units (GRU) and transformer-based models (AraBERT v1, AraBERT v02, AraBERT v2, ArElectra, QARiB, Arbert, and Marbert). They obtained an F1 score, Precision, Recall and Accuracy of 67%, 96%, 54% and 53%, respectively.

Due to the small number of studies addressing fake news detection in Arabic, we also revised prior work in other languages such as articles [14], [15], [16] and [17] who discussed English fake news detection using different techniques. In [27], while examining English fake news detection, the authors proposed a two-step approach for detecting fake news in social media. In the first step of the method, several pre-processing stages were applied to the dataset to convert un-structured data into a structured dataset. This dataset contained news represented as vectors. Their second step was applying twenty-three supervised artificial intelligence algorithms to the structured dataset using text mining methods. With three different real-world datasets, they used the decision tree machine-learning algorithms ZeroR, CVPS, and WIHW. The maximum accuracy they achieved was 74.5%, with precision recorded at 74.1%, recall at 78% and F-score at 75.9%. This method is expected to achieve better performance when using Deep Learning models rather than classical machine learning.

Traylor et al [28], developed a novel NLP model to identify English fake news. The Text Blob, Natural Language, and SciPy Toolkits were used to develop a novel fake news detector that uses quote attribution in a Bayesian machine learning system to estimate the likelihood that a news article is false. They concluded the paper by stating how this work would evolve into an influence mining system. However, this work did not achieve a powerful performance, the researchers measured the model's accuracy at 69.4%.

Using BERT model, Sadeghi et al [29] described the entry of the Intelligent Knowledge Management (IKM) Lab in the WSDM 2019 Fake News Classification challenge. They treated the problem as a natural language inference problem. They achieved an accuracy of 88.63%.

Compared with other papers, authors in [30] proposed a strong work in the field of English fake news detection. The authors introduced their dataset, which was composed of several merged datasets collected from Kaggle and other resources, with a total of more than 6000 instances. They also used various classical machine learning and deep learning models such NB, LSTM, CNN, random forest, KNN and very deep CNN. They explored the benefits of feature extraction and how it can affect the model. When they implemented a cascaded model containing CNN and LSTM, they recorded a performance of 97.3%, while the very deep CNN achieved 98.3%. The authors in [31] addressed the current methods for identifying fake news, news domains and Twitter bots. The new model relies on advances in Natural Language Understanding (NLU) end to end deep-learning models to identify stylistic differences between legitimate and fake news articles. The other model identified the domain of each news instance such as politics, sports, religion, etc. For Twitter bot detection, they used distinctive features to determine the reliability of the poster such as tweet time, duration between account creation and tweet date, user's location, and other features. They used five models with different parameters. To detect bots, they targeted Arabic tweets through hashtags, while for fake news and domain detection, they targeted six English datasets. They evaluated their performance using precision, recall, and F-score, which were recorded at 92%, 100% and 96% for mBert base and, 98%, 98% and 98% for XLNET, respectively.

We see from the related work that there is a lack of Arabic datasets constructed specifically for fake news detection tasks. In addition, detection model performance for Arabic text is poor for such an important system. Truly effective models should not perform at lower than 95% accuracy, as there is a chance that the model will label fake news as true or true news as fake, resulting in a credibility issue. More than 80% of the listed papers used classical machine learning techniques, while only 20% of them used deep learning or transformers. Our study produces strong results in terms of accuracy when compared to related work in Arabic and English.

### III. METHODOLOGY

In the following figures, we summarize the methodology we used to build our fake new detection model. As a first step we prepare the datasets we are going to use in the experiments, and they are two. For the original Arabic dataset, we collected it using web scraping from different resources. Whereas we translate an English dataset using Google API. We process both datasets, filter them and feed them to the tokenizer. At the end of this step, we have two ready datasets as shown in Figure 1.

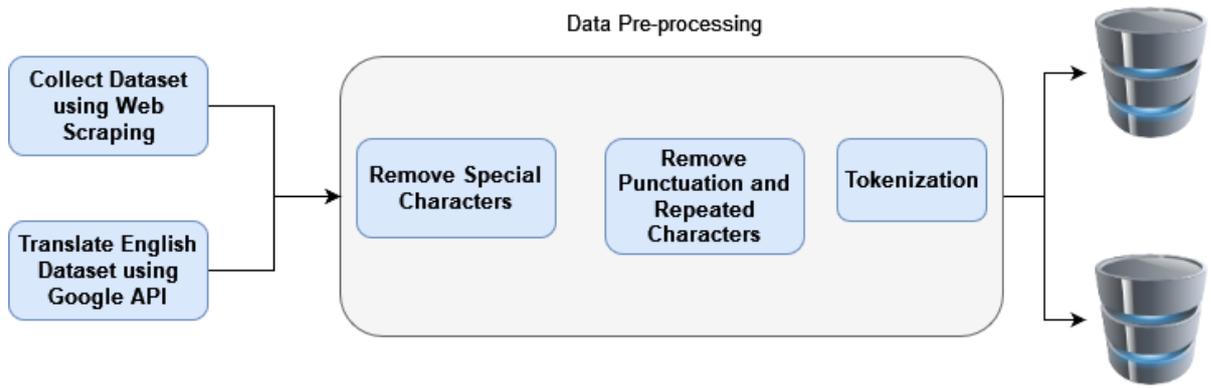

*Figure 1 Data preprocessing methodology*

For our fake news classification task, we chose eight recently developed Arabic contextualized embedding models. As shown in the figures below, the pre-trained models have already been trained on a large amount of Arabic data. As a result, these models are familiar with the Arabic language, letters, vocabulary and stop words, as well as how words are used in a sentence. Our contribution is to fine-tune these BERT models for our task, which is to train them on our filtered data with the goal of reliably classifying Arabic fake news.

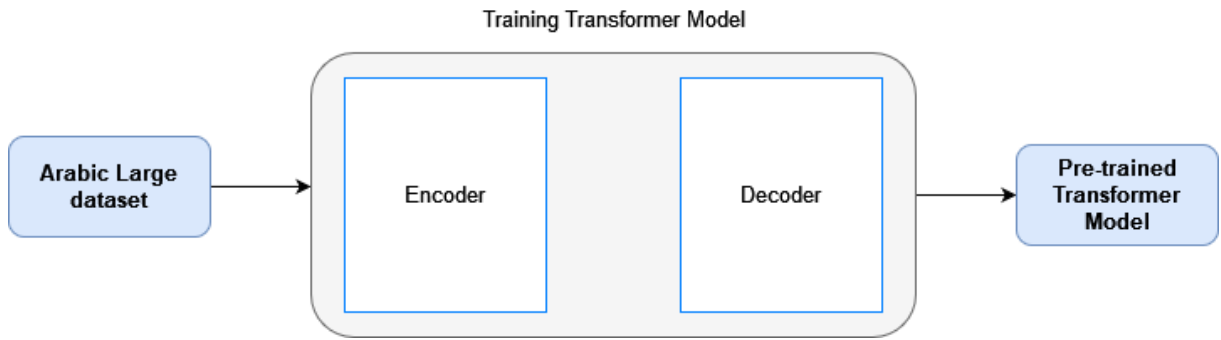

*Figure 2 Pre-trained Transformer model*

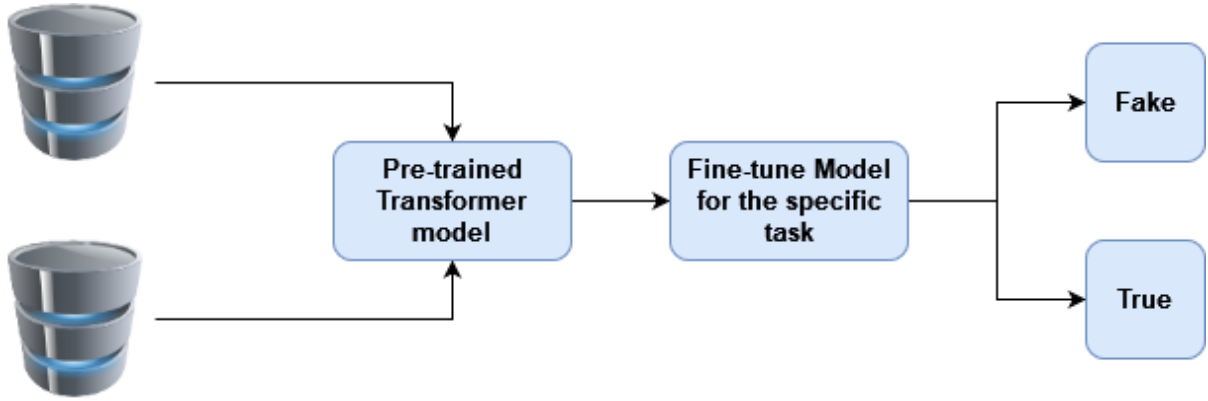

Figure 3 Fine-tune and Train Fake news detection model

A. *Models and technicals*

Deep learning is a field of machine learning that deals with different types of unstructured data such as images, text data and speech data. Deep learning algorithms also conduct feature extraction for the purposes of classification, unlike traditional machine learning, which requires manual feature extraction in order to proceed to classification tasks. Deep learning has been developed to achieve excellent performance in several fields such as speech recognition, image processing, medicine, finance, and NLP.

In this research, we are implementing one of the most important NLP tasks. NLP is a field of artificial intelligence that deals with human text and language. NLP can be used for many tasks, including translation from one language to another and supplying automated chat box services.

In the last years, transformers have been introduced and used at an increasing rate. Transformers are deep learning models released in 2017 and mainly used in NLP tasks. Later in this section, we will introduce some of the transformer models used in this paper such as Bert, GigaBERTv4 - base, Arabert and Arabic-Bert.

The below eight models was chosen to conduct our experiments based on their ability to perform NLP tasks for Arabic text or multilingual text based on the reviewed literature work. We used XLM-Roberta and GigaBERTv4 as those two models are multilingual, trained on many languages including Arabic. Whereas we used Arabert, Arabic-Bert, ArBert, MARBert, Araelectra and QaribBert – base as they are trained on Arabic text using Bert structure. The models show good performance in different NLP tasks for Arabic text in the literature work in which we can conclude that they have the ability to outperform the normal Bert when it comes to Arabic language.

BERT stands for Bidirectional Encoder Representations from Transformers. BERT models are NLP models that have been trained on a large amount of data. Instead of using the traditional techniques of work tokenization, they deal with the meanings of words. They can also learn sentences either from left to right or from right to left.

This makes them reliable for use in many languages such as English, Arabic, Hindi, etc.[5] BERT models are documented in detail on Hugging Face.[6] This platform provides a detailed guide to using any BERT model for different applications. BERT models can also be used with Pytorch and TensorFlow.

XLM-Roberta is a multilingual model trained on over 100 languages. The main advantage of this model is that there is no need to mention which language is being used, unlike other Multilingual models that need to provide a language tensor. It recognizes the language automatically using the input IDs generated by the tokenizer [32].

GigaBERTv4 base is a BERT model customized for bilingual BERT in both Arabic and English.[7] According to the documentation, it was trained on a large amount of data (Giga word + Oscar + Wikipedia) with more than 10 billion tokens, showing zero shot performance from English to Arabic on information extraction [33]. Zero-shot learning is a problem setup in machine learning. At test time, the learner observes samples from classes that were not observed during training and is required to predict the category they belong to[8].

The Arabert[9] model, provided by AUB University, is a BERT architecture based on the Arabic language. In this model, the authors tried to achieve the same success that BERT did for the English language for most NLP tasks. It was released in March 2020 and it has already achieved optimal performance on most assigned Arabic NLP tasks [34]. Various studies on this Arabert architecture explain the training methodology in detail. It is worth mentioning that the model has been pre-trained on a large dataset consisting of 70 million sentences, corresponding to ~24GB of text [34].

AUB also released the model AraELECTRA [10], a BERT tool designed for Arabic text. Like Arabert, AraELECTRA depends on a BERT structure and has achieved new standards in Arabic question answering. Two version of this model were released: the AraELECTRA-base-discriminator and the AraELECTRA-base-generator. In this study, we used the discriminator because the structure is stronger and has more hidden size and attention heads.

Another model we use is QaribBert – base. This model was trained on around 420 million Tweets and 180 million sentences of text, all of them in Arabic, collected using a Twitter API [35].

---

[5] https://www.analyticsvidhya.com/
[6] https://huggingface.co/
[7] https://huggingface.co/
[8] https://en.wikipedia.org/wiki/Zero-shot_learning
[9] https://github.com/KUIS-AI-Lab/Arabic-ALBERT
[10] https://arxiv.org/pdf/2012.15516.pdf.

Moreover, we also used two more powerful transformer-based language models for Arabic called ARBERT and MARBERT11, released by UBC. These two models were tested on five tasks for the Arabic language and each produced positive results. Like the other models in this study, they use BERT-based architecture: 12 attention layers, each of which has 12 attention heads and 768 hidden dimensions. In this study, the ARBERT model produced very reliable results.

Finally, we used the Arabic-BERT model provided by Ali Safaya in GitHub12. The model was pre-trained on approximately 8.2 billion words. The advantage of this model is that the final version of the training dataset contains some non-Arabic words that were not removed from the texts. This may affect fake news detection tasks in a positive way [36].

## B. *Dataset*

For comparing purposes, we used two datasets in our experiments, translated data and originally Arabic data.
### 1) Translated data

In the first experiment, we include a fake news detection dataset from Kaggle[13] to build our classification model. The data is originally written in English, but since we are conducting Arabic text-based fake news detection, we need to translate the dataset into Arabic. Originally the dataset consists of five columns: ID, title, author, text, and the label column that marks the article as potentially unreliable, where 1 is used for unreliable/fake and 0 for reliable/real. We translated the titles to Arabic using python code and the package "google-trans", after which we exported it into a csv format.

The data contains some Russian and German letters that we could not translate using our English to Arabic translator code, so we removed them as the first step in the pre-processing technique. After that, we removed the remaining English words that could not be translated to Arabic such as names and terms (ex: Donald Trump, Jennifer, etc.).

To begin with, we translated a balanced dataset consisting of only 10,000 instances, after which we increased the pool by 10,000. The Kaggle dataset originally contained 20,000 instances, but after we pre-processed the data, we ended up with 16,000 instances in the data pool. We removed punctuation and unnecessary stopping words, as well as English and Arabic numbers. Finally, during the training process, we tokenized each sentence using the pre-trained model tokenizer. According to the figure below, around 32% are fake news and 68% are real, which indicates that the dataset is not balanced.

---

[11] https://github.com/UBC-NLP/marbert#4-how-to-use-arbert-and-marbert

[12] https://github.com/alisafaya/Arabic-BERT

[13] https://www.kaggle.com/c/fake-news/data?select=test.csv

*2) Collected data*

In this step, we constructed our Arabic dataset. Finding an Arabic fake news dataset was not an easy job, as even the pre-made datasets did not contain enough data. We collected fake news from two resources. We obtained 4000 news sentences from an Arabic Twitter dataset provided by Mendeley[14] that contains rumor tweets originally written in Arabic. Another 1000 data instances were obtained using web scraping from the well-known No Rumors website archive.[15]

We obtained the true news from trusted and official news sites. For example, 4000 instances were sourced from al-Arabiya[16] and 1000 more from al-Khaleej.[17] Both datasets are concerned with political issues related to the Middle East. Our dataset is balanced, with 5000 from each class, which leave us with a total of 10,000 instances in our dataset. According to the figure below, around 49% of the data was found to be fake news, while 51% was real data, which indicates a fairly balanced dataset. For the fake data from the Twitter dataset, we removed special characters such as emojis, punctuation and repeated characters as a step in the pre-processing stage.

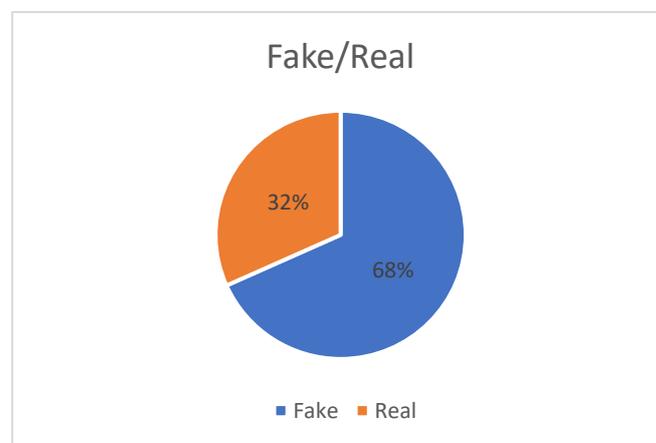

*Figure 4 Translated data fake and true data frequency*

---

[14] https://data.mendeley.com/datasets/9sht4t6cpf/2
[15] http://norumors.net/?post_type=rumors?post_type=rumors
[16] https://english.alarabiya.net/#slide=4
[17] khaleejtimes.com

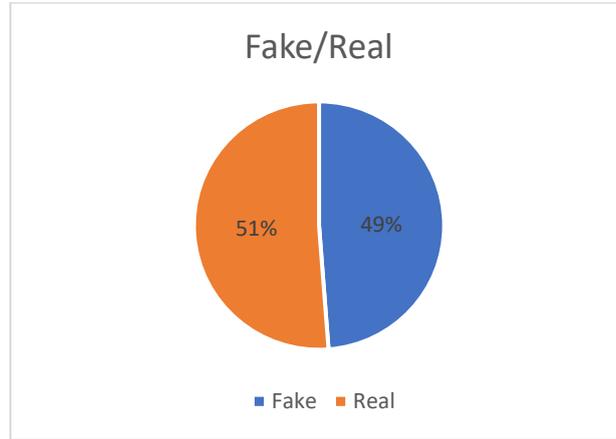

*Figure 5 Original collected Arabic data fake and true data frequency*

The following table shows how our code pre-processes the data before filtering it and entering it into the tokenizer.

*Table 1 Before and after pre-processing*

| Sentence before pre-processing | Sentence after pre-processing |
|---|---|
| 👍👍 المملكة تبرم اتفاقية مع وفد الفاتيكان بقيادة الكاردينال توران لفتح كنائس للمسيحين المقيمين في السعودية | المملكة تبرم اتفاقية مع وفد الفاتيكان بقيادة الكاردينال توران لفتح كنائس للمسيحين المقيمين في السعودية<br><br>The Kingdom signs an agreement with the Vatican delegation led by Cardinal Tauran to open churches for Christians residing in Saudi Arabia |
| طلع عندنا تنيل (تنين) في اليمن وفِي ابين مسقط راس والدي ❤️??? انتو شو طلع عندكم | طلع عندنا تنيل تنين في اليمن وفِي ابين مسقط راس والدي  انتو شو طلع عندكم<br><br>We had a dragon in Yemen and in Abyan, my father's hometown. What about you? |
| 1هبوط طائرة بين مكة و جدة على الخط السريع?????✈️? | هبوط طائرة بين مكة و جدة على الخط السريع<br><br>A plane lands between Mecca and Jeddah on the highway |
| السعودية تقرر منع تشغيل الموسيقى في المطاعم و المقاهي بـ الرياض ⬅️ eXtranews | السعودية تقرر منع تشغيل الموسيقى في المطاعم و المقاهي ب الرياض<br><br>Saudi Arabia decides to ban playing music in restaurants and cafes in Riyadh |
| حصرياااااا اساب الانفجار الذي حدث في جدة ومن الفاعل شاهد التفاصيل بالصور : | Exclusively because of the explosion that occurred in Jeddah and the perpetrator. See the details in the pictures |

*C.* **Configuration**

In the configuration stage, we added nine deep learning layers on top of the pre-trained structure of each model. We added a linear layer followed by a dropout layer and a ReLU layer. We repeated this three times, then we added a SoftMax layer with one dimension to end up with a total of 3 linear layers, 3 dropout layers, 2 ReLU layers and one SoftMax layer.

The linear layer we added is to find a strong correlation between the input text and the output labels, the dropout will improve the model fitting resulting in increased accuracy. The ReLU layer will decrease the computing time for the model training.

The following diagram summarizes our algorithm:

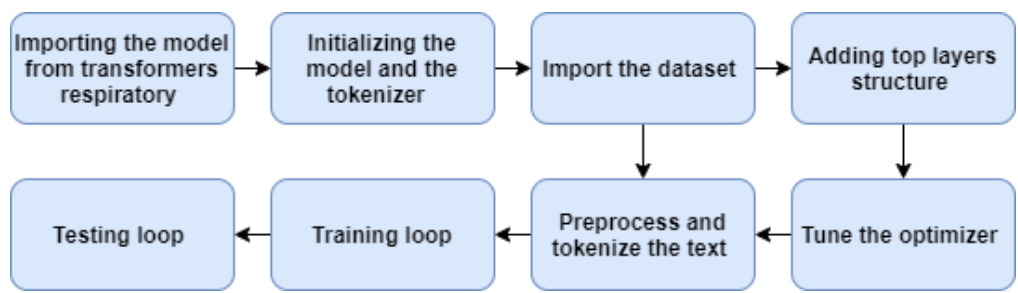

*Figure 6 Configuration Summary*

Our evaluation process was based on many parameters in the training and the testing phases. In the training phase, we analyzed the loss value in every step of the training to see when it started to decrease. Loss is a number evaluating the effectiveness of each model's prediction. The closer the value is to zero, the more accurate the model's predictions.

## IV. EXPERIMENTS

In this section we describe our fake news classification system. First, we describe the fine-tuning and optimization that we used to tune our models to get the highest performance. Next, we list the evaluation criteria that we used to assess our models during both the training and the testing phases. We describe the outcomes and results for our first experiment, which was conducted using the translated dataset from Kaggle. We then lay out the results for the second experiment using the collected dataset. Finally, we discuss the general results and sum up our findings.

A. EXPERIMENTAL FINE-TUNING

A BERT model is a pre-trained model that is already built and trained on a large amount of data. When we use it in a specific task, we need to fine-tune the parameters. During the fine-tuning process, we can adjust many parameters such as the optimizer, learning rate, number of epochs and the dropout value. Here are the parameter values we used:

- Optimizer: We tried more than one optimizer as a part of the fine-tuning process. We tried the SGD optimizer, ADAM and finally ADAMW.
- Learning rate: as we get many well-known values from previous fake news detection codes for English text, we tried 0.001, 0.0001 and finally 1e^-5.
- Number of epochs: we tried many numbers of epochs from 1 to 100. We applied the stop accuracy method that allowed us to stop the training when we reached the highest accuracy, so as not to waste time and storage.
- Dropout value: we tried 0.1, 0.25 and 0.5, with each value producing a slightly different result.

In the last step, we divided our data randomly into 80% training and 20% testing.

B. PERFORMANCE EVALUATION

Our evaluation process depended on many parameters in the training and the testing phases. In the training, we assessed the loss value in every step of the training to see when it started to decrease. Loss is a number evaluating the effectiveness of each model's prediction. The closer the value is to zero, the more accurate the model's predictions.

In the testing phase, we labeled a task as **TP** (true positive) when the model classifies a given news instance as true and it is actually true [37] and **TN** (true negative) when the instance is fake and the model labels it as fake. **FP** (false positive) was the label given when the model falsely predicts the instance to be true and **FN** (false negative) is when the instance is actually true, but the model classifies it as fake. We put each in a file to observe them later.

We then calculated each model's overall accuracy to better understand its performance:

$$Accuracy = \frac{TP + TN}{TP + FP + TN + FN}$$

We obtained each model's precision, recall and F-score using the following formulas:

$$precision = \frac{TP}{TP + FP}$$
$$recall = \frac{TP}{TP + FN}$$
$$F1 = \frac{2 \times precision \times recall}{precision + recall}$$

We built our initial fake news classification system using dataset A. As we fine-tuned the system, we tried many configurations and parameters to get the best possible results.

Table 2 shows all the trials we did along with the corresponding results for all the models, so the best values are visible for each variable.

In addition to the evaluation, we plotted the loss values during the training process to observe how the process developed.

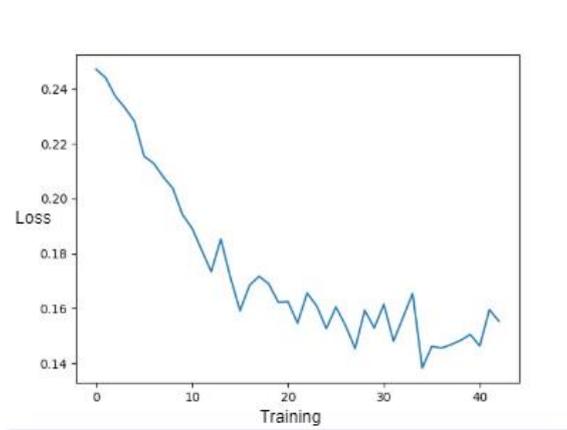

*Figure 7 Loss values for translated data*

From the data, we can see that the loss values decrease during the training process, which is a positive development. However, the minimum loss value the model reached was around 0.15, which is still too high for a fake news detection system. It is extremely important that such a system be reliable, as the effects of false negatives and false positives can have detrimental effects on society.

*Table 2: Model performance using translated data with fine-tuning steps.*

| Model | Accuracy | Precision | Recall | F-score |
|---|---|---|---|---|
| GigaBert-base | **91.2%** | **88.5%** | **95.3%** | **91.8%** |
| RobertaBase | 75.9% | 84.9% | 64.5% | 73.3% |
| Arabert | 81.9% | 79.2% | 87.5% | 83.1% |
| Arabic-Bert | 89.4% | 85.8% | 94.9% | 90.1% |
| ArBert | 84.2% | 82.1% | 88.4% | 85.1% |
| MARBert | 81.2% | 77.0% | 90.2% | 83.0% |
| Araelectra | 78.5% | 79.4% | 78.1% | 78.8% |
| QaribBert – base | 85.7% | 80.8% | 94.4% | 87.1% |

To obtain the data in the above table, all models were set to 25 epochs using a dataset size of 16,000 instances. To further fine-tune the best performing classifier (GigaBert-base), we re-trained it using 100 epochs. The classifier achieved an accuracy score of 93.6%. It was clear that the performance improved when we increased the number of epochs.

In addition, we noticed that the high dropout value was detrimental to the model's performance, so we opted to use the structure without dropout layers. The system performed with the highest accuracy when we used the GigaBert algorithm without dropout layers. However, even in those conditions, its performance was not considered sufficiently accurate to be used as a fake news detection system. When we looked at the false negative and false positive files in an attempt to establish where and why the model's predictions are inaccurate, we see that in these cases, the model cannot find keywords to differentiate between fake and true sentences. This is because there are many words in the translated data that are not found in the Arabert and the BERT-Arabic based vocabulary files, such as ['هيلاري', 'بيرس مورجان', 'نيويورك تايمز', 'دونالد ترامب', 'هابلیس']. These are the types of words that ended up producing false negatives.

On the other hand, when we were using Arabert and Arabic-BERT, the two models that were originally trained on Arabic data from various sources, the vocabulary files for these models produced consistent results in a way that was not possible in the translated data. For example, see the following sentence from the translated dataset:

['بيرس مورجان: "انزل عن حصانتك العالية ، هيلاري. مرشحة واحدة فقط تصل إلى رقبتها في تحقيقات مكتب التحقيقات الفيدرالي واسمها ليس ' دونالد -TruthFeed"]

You can see the words 'هيلاري' and 'بيرس مورجان', which are not found in the Arabic-based BERT vocabulary file. As a result, we collected another dataset that was originally written in Arabic and that contained the words that exist in the vocabulary files of the model.

In this experiment, we depended on the 'stop' accuracy method that leaves us always with 10 epochs. After many trials, we decided to use the AdamW optimizer with no dropout value.

We also logged the loss value updates during the training process. The following figure represents the training loss values for the Arabic-BERT model.

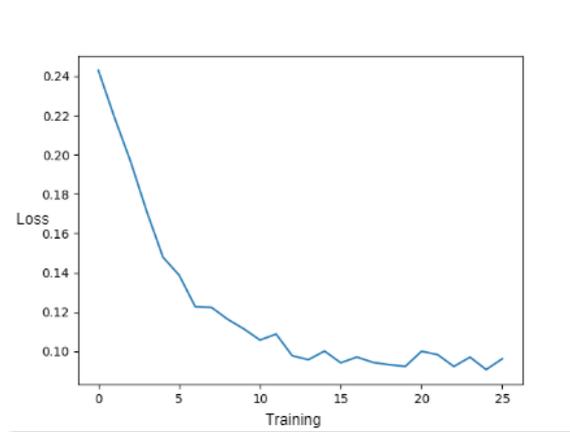

*Figure 8 Loss values recorded for the collected data*

In the above figure, we can see that the loss value decreased very quickly, and that it remained stable after dropping.

*Table 3: Model performance using the collected data.*

| Model | Accuracy | Precision | Recall | F-score |
|---|---|---|---|---|
| GigaBert - base | 97.4% | 96.8% | 98.6% | 97.7% |
| Roberta-Base | 85.6% | 82.4% | 94.2% | 87.9% |
| Arabert | 96.5% | 96.5% | 97.3% | 96.3% |
| Arabic-BERT | 98.0% | 99.1% | 97.0% | 98.1% |
| ARBERT | 98.8% | 99.4% | 98.3% | 98.9% |
| MARBERT | 96.9% | 96.2% | 98.3% | 97.2% |
| Araelectra | 80.0% | 79.6% | 86.7% | 83.0% |
| QaribBert – Base | 98.5% | 99.1% | 98.2% | 98.6% |

In this experiment, we used the dataset we constructed out of 10,000 samples. We can see that the results obtained using this new data is much better than those obtained using the translated data. In fact, the best performance was recorded at 98% accuracy using the Arabic-BERT, ARBERT and QaribBert pre-trained models.

The following table compares our results with that of previous works.

*Table 4: Comparison with similar works for Arabic-language data.*

| Paper | Model | Accuracy | Precision | Recall | F-score | Dataset size | Balanced | Translated/Arabic |
|---|---|---|---|---|---|---|---|---|
| [1] | No Embedding | 74.0% | 78.0% | 80.0% | 79.0% | 2K | No | Arabic |
| [5] | Arabert | 72.5% | - | - | 62.0% | 7K | No | Arabic |
| Our work | ARBERT | 98.8% | 99.4% | 988.3% | 98.9% | 10K | Yes | Arabic |
| Our work | Arabic-BERT | 98.0% | 99.1% | 97.0% | 98.1% | 10K | Yes | Arabic |
| Our work | QaribBert – Base | 98.5% | 99.1% | 98.2% | 98.6% | 10K | Yes | Arabic |

*C. Discussion*

From our study, we can conclude that when we use the dataset that we constructed from Arabic-language news instances, we obtain the best performance. This is because our models were originally trained in Arabic vocabulary using various Arabic-language resources including Arabic Wikipedia. Giga-Bert also performed relatively well in terms of Arabic text. In fact, when compared to recent work published about fake news detection in Arabic, our results are superior to all other work when using an Arabic dataset. The next best performance was produced by

[38], which obtained 87.2% accuracy and an 89.21% F-score, whereas we achieved 96.5% accuracy and a 96.9% F-score using Arabert.

To improve our model, we examined the false positives and the false negatives, but we did not find any common features among the sentences. Table 5 contains some examples of the sentences that the model classified incorrectly.

*Table 5: Samples of false negatives and false positives produced by our model*

| False negatives | False positives |
|---|---|
| '[هز انفجار عنيف مدينة عدن]' | '[شارك الآلاف في احتجاج وسط لندن أمس الأول السبت، ضد تجديد نظام ترايدنت البريطاني للأسلحة النووية]' |
| '[بايع الأمير محمد بن نايف بن عبدالعزيز الأربعاء الأمير محمد بن سلمان بن عبد العزيز ولياً للعهد في قصر الصفا بمكة]' | '[للمرة الأولى التحالف العربي يحذر العالم من خطر الحوثيين والأمم المتحدة تزف بشرى لليمنيين ]' |
| '[قبل عامين ونصف العام اختفى خبير تكنولوجيا المعلومات اللبناني نزارزكا في إيران]' | '[صدق أو لا تصدق العثور على تنين في محافظة أبْين في اليمن]' |
| '[الأسواق الدمشقية العتيقة رونق الشام وطابعها]' | '[وضع علم المملكة العربية السعودية على قوس النصر في باري]' |
| '[عمليات الاجلاء متوقفة في حلب منذ الجمعة]' | '[نعوم تشومسكي يتحدث عن توقعاته حيال علاقة مصر الاخوان و واشنطن بعد الثورة المصرية حاوره]' |
| '[فاز الرئيس عبد العزيز بوتفليقة بولاية رابعة بنسبة]' | '[تضارب الأنباء بشأن وفاة الرئيس المصري السابق حسني مبارك سريريا]' |
| '[ما زالت أخبار فضيحة تمويل قطر لتنظيم إرهابي في العراق تتدفق]' | '[زيارة الرئيس الصيني للمساجد وطلبه من المسلمين أن يدعوا الله لرفع ب]' |

Our best-performing model classified 35 fake news instances as true (FP) out of the complete dataset of 10,000 instances, and 129 true news instances as fake (FN). When we discuss fake news detection systems, the instances of false negatives and false positives are important. When a given article is fake and the model predicts it to be true, it will spread untrustworthy news, producing negative consequences in society. On the other hand, if the model falsely predicts a true news instance to be fake, users may stop trusting the site, producing an unjustifiable lack of confidence in the public. In either case, a false result will have negative consequences that may cause harm and lead to a real problem.

To conclude this discussion, we can say that in case of fake news detection, the potential harm caused by false positive and false negative results is relative, which means that it depends on the given scenario. For instance, if we have a news instance that claims that a given brand of Covid-19 vaccine is effective when in fact that vaccine is harmful to humans, and the system classifies that news as true, that false positive label will have dire consequences on society. However, if a news instance claims that wearing face masks protect people from Covid-19, which is true, but the system labels it as false, this will make the disease spread in a very quick and catastrophic way. Fake news detection is not like spam detection, where a false positive (falsely labeling a message as spam) is much worse than a false negative (falsely labeling a spam message as trustworthy). Unlike machine learning techniques, the deep learning contextualized embedding models have feature extraction phase to classify sentences. Examining the true negative and true positive instances, we noted some common keywords that the model may be using to distinguish between fake and true news instances. In Table 5, we list some of these keywords.

*Table 6: Keywords observed in true predicted instances*

| TP | TN |
| --- | --- |
| أكدت | انفجار |
| سوريا | هام |
| داعش | عاجل |
|  | اربح |
|  | جوائز |

These keywords may vary when we change the date of the news in the dataset knowing that our dataset is up-to-date, and the news is very recent.

## V. CONCLUSIONS

In this paper, we developed several transformer models while utilizing a variety of contextualized Arabic embedding models for the purpose of fake news detection in Arabic text. Using two new datasets, which were constructed during this work, we were able to achieve excellent performance for the fake news detection task. Our study consisted of eight transformer models, five of which had never been used before for fake news classification. The two types of data we used were data that was translated from English and data that we collected from Arabic websites and Arabic tweets. The models that were trained on the collected Arabic data produced better results than those trained on translated data. We collected the dataset that was originally written in Arabic, getting the misleading sentences from Twitter and the reliable ones from official news sites such as Al-Arabiya and al Khaleej. We then annotated them as fake or true. The fake news identification task was performed using transformers' architecture utilizing state-of-the-art contextualized Arabic embedding models. These models are Giga-Bert, Roberta-Base, AraBert, Arabic-BERT, ARBERT, MarBert, Araelectra and QaribBert. The performance evaluation process showed that Roberta-Base produced the lowest performance for both datasets, while ARBERT and Arabic-Bert performed the best, with 98.8% and 98% accuracy, respectively. These are some of the highest accuracy scores for an Arabic fake news detection task ever reported. For future work, we will collect more data in the form of tweets and identify the fake ones. We will also collect fake data and rumors from anti-rumor sites such as No Rumors and obtain reliable data from official news sites. As the dataset increases in size, the performance of the models will improve. In this world that relies so heavily on social media, it has become very difficult to track all posts on Twitter and other platforms. Fake news detection systems are therefore important and will provide an important contribution to addressing these issues.

**Compliance with Ethical Standards:** The authors would like to convey their thanks and appreciation to the "University of Sharjah" for supporting the work through the research group – Machine Learning and Arabic Language Processing

**Conflict of Interest:** The authors declare that they have no conflict of interest.

**Informed consent:** This study does not involve any experiments on animals.